\documentclass[lettersize,journal]{IEEEtran}

\usepackage{amsmath,amsfonts}
\usepackage{algorithm}
\usepackage{algcompatible}
\usepackage{array}
\usepackage[caption=false,font=normalsize,labelfont=sf,textfont=sf]{subfig}
\usepackage{amsthm} 
\usepackage{textcomp}
\usepackage{stfloats}
\usepackage{url}
\usepackage{verbatim}
\usepackage{graphicx}
\usepackage{cite}

\usepackage{xcolor}

\usepackage{caption}   
\allowdisplaybreaks

\usepackage{annotate-equations}

\newcommand{\nfeatures}{\ensuremath{p}}
\newcommand{\nrank}{\ensuremath{k}}
\newcommand{\nsamples}{\ensuremath{n}}

\newcommand{\realSpace}{\ensuremath{\mathbb{R}}}
\newcommand{\Rplusplus}{\ensuremath{\realSpace_{++}}}
\newcommand{\oblique}{\ensuremath{\mathcal{OB}}}
\newcommand{\orthGroup}{\ensuremath{\mathcal{O}}}
\newcommand{\SPSD}{\ensuremath{\mathcal{S}^{+}}}

\newcommand{\obliqueTangent}[1]{\ensuremath{T_{#1}\mathcal{OB}}}
\newcommand{\horizontal}[1]{\ensuremath{\mathcal{H}_{#1}}}
\newcommand{\vertical}[1]{\ensuremath{\mathcal{V}_{#1}}}

\newcommand{\metric}[3]{\ensuremath{\langle #2, #3 \rangle_{#1}}}

\newcommand{\eye}{\ensuremath{\mathbf{I}}}
\newcommand{\zero}{\ensuremath{\mathbf{0}}}

\newcommand{\data}{\ensuremath{\mathbf{x}}}
\newcommand{\dataMat}{\ensuremath{\mathbf{X}}}
\newcommand{\SCM}{\ensuremath{\mathbf{S}}}

\newcommand{\precision}{\ensuremath{\boldsymbol{\Theta}}}

\newcommand{\powers}{\ensuremath{\boldsymbol{\sigma}}}
\newcommand{\powersTangent}{\ensuremath{\boldsymbol{\xi}_{\boldsymbol{\sigma}}}}
\newcommand{\powersTangentBis}{\ensuremath{\boldsymbol{\eta}_{\boldsymbol{\sigma}}}}

\newcommand{\W}{\ensuremath{\mathbf{W}}}
\newcommand{\Wtangent}{\ensuremath{\boldsymbol{\xi}_{\mathbf{W}}}}
\newcommand{\WtangentBis}{\ensuremath{\boldsymbol{\eta}_{\mathbf{W}}}}

\DeclareMathOperator{\D}{D}
\DeclareMathOperator{\grad}{\nabla}
\DeclareMathOperator{\ddiag}{ddiag}
\DeclareMathOperator{\diag}{diag}
\DeclareMathOperator{\tr}{tr}

\DeclareMathOperator{\rank}{rank}

\newtheorem{remark}{Remark}

\title{
Leveraging Low-rank Factorizations of Conditional Correlation Matrices in Graph Learning
}

\author{
Thu Ha Phi, \IEEEmembership{Student Member, IEEE}, Alexandre Hippert-Ferrer, Florent Bouchard, Arnaud Breloy
\thanks{Thu Ha Phi is with LEME (EA 4416), Université Paris Nanterre, France, email: phi.thuha@parisnanterre.fr}%
\thanks{Alexandre Hippert-Ferrer is with LaSTIG, IGN ENSG, Université Gustave Eiffel, France, email: alexandre.hippert-ferrer@ign.fr}%
\thanks{Florent Bouchard is with L2s, Université Paris Saclay, CNRS, CentraleSupélec, France, email: florent.bouchard@centralesupelec.fr}
\thanks{Arnaud Breloy is with CEDRIC (EA 4629), Conservatoire National des Arts et Metiers, email: arnaud.breloy@cnam.fr}
\thanks{This work was supported by the MASSILIA project (ANR- 21-CE23-0038-01) of the French National Research Agency (ANR).}
}

\begin{document}

\maketitle

\begin{abstract}
This paper addresses the problem of learning an undirected graph from data gathered at each nodes.
Within the graph signal processing framework, the topology of such graph can be linked to the support of the conditional correlation matrix of the data.
The corresponding graph learning problem then scales to the squares of the number of variables (nodes), which is usually problematic at large dimension.
To tackle this issue, we propose a graph learning framework that leverages a low-rank factorization of the conditional correlation matrix.
In order to solve for the resulting optimization problems, we derive tools required to apply Riemannian optimization techniques for this particular structure.
The proposal is then particularized to a low-rank constrained counterpart of the GLasso algorithm, i.e., the penalized maximum likelihood estimation of a Gaussian graphical model.
Experiments on synthetic and real data evidence that a very efficient dimension-versus-performance trade-off can be achieved with this approach.
\end{abstract}

\begin{IEEEkeywords}
Graph learning, low-rank models, Riemannian Optimization, unsupervised learning.
\end{IEEEkeywords}

\section{Introduction}
\IEEEPARstart{G}{raphs} provide fundamental structures in order to represent complex networks: nodes represent entities, and edges define the relationships between them.
Graph thus emerged as ubiquitous tools for data visualization and exploratory analysis across various domains, including social networks, biological networks, transportation systems, financial engineering, and sensor networks \cite{ortega2018graph, dong2020graph, leus2023graph}.
Moreover, the knowledge of the graph topology also provides valuable insights for various tasks, such as node classification (assigning each node to an appropriate category), graph clustering (grouping similar nodes into the same category), graph signal processing and graph neural networks (processing data at each nodes). 
Such topology is not always known in practice, which has motivated extensive research in \textit{graph learning}, where the primary objective is to infer the graph from the data generated at each node \cite{friedman2008sparse, egilmez2017graph, zhao2019optimization, kumar2020unified, ying2020nonconvex, hippert2023learning, phi2024leveraging, medvedovsky2024efficient,buciulea2025polynomial, mazumder2012graphical, lauritzen2001causal, lauritzen2018unifying, fallat2017total, Lauritzen2017MaximumLE, cai2023fast, vogel2011elliptical, liu2012transelliptical}. 

In this scope, many works related to graph signal processing build upon a statistical framework in which nodes represent variables, and where an edge represents a conditional correlation relationship.
Formally, a graph of $p$ nodes produces samples that are data vectors, and an edge between the nodes $q,\ell \in \! [\![1,p]\!]^2 $ exists if the conditional relation between the corresponding variables is non-zero.
A graph learning problem consists then in estimating the underlying structure of conditional correlation from a sample set of data gathered at all edges.
Since this structure can be linked to the support of the precision matrix in many multivariate statistical models \cite{lauritzen2001causal, lauritzen2018unifying}, graph learning is then commonly framed as a structured covariance (or precision) matrix estimation problem.
Within this paradigm, Gaussian graphical models (GGM), also referred to as Gaussian random Markov field (GRMF), assume that the data is sampled according to a multivariate Gaussian distribution of covariance matrix $\boldsymbol{\Theta}$, and corresponding precision matrix $\boldsymbol{\Theta}=\boldsymbol{\Sigma}^{-1}$.
The reasonable assumption of a graph with only few edges per nodes translates then in a sparse structure on the precision matrix.
A prominent graph learning method in this context is brought by the Graphical Lasso (GLasso) algorithm \cite{friedman2008sparse, mazumder2012graphical}. 
This approach formulates the graph learning problem as a penalized maximum likelihood estimation of the precision matrix of GGMs, in which the $\ell_1$-norm is used as penalty to promote a sparse structure on its off-diagonal elements.

Following from GLasso, subsequent research has proposed model refinements in several directions. 
A first direction addressed the integration of additional prior structure knowledge, allowing for a better fit to the data at hand, thus achieving a better graph estimation at low sample support.
This integration can be performed by adding constraints over the covariance matrix $\boldsymbol{\Sigma}$ or the precision matrix $\boldsymbol{\Theta}$.
In this setup, many statistical models were explored: \cite{fallat2017total, Lauritzen2017MaximumLE, cai2023fast} investigated the total positivity of order two (non-negative conditional correlations), \cite{hippert2023learning} proposed algorithms for factor models, while \cite{phi2024leveraging} introduced standardization-based constraints.
Additional structures also emerged from the graph signal processing framework, that allows for linking samples to a particular model involving graph shift operators \cite{ortega2018graph, dong2020graph, leus2023graph}.
A popular model tied to the notion of smooth signals over a graph enforces $\boldsymbol{\Theta}$ to be a Laplacian matrix in GGMs (also referred to as Laplacian constrained Gaussian random Markov fields) \cite{egilmez2017graph, ying2020nonconvex, medvedovsky2024efficient}, and possibly constraining its eigenstructure \cite{zhao2019optimization, kumar2020unified}.
A second direction relates to the relaxation of the Gaussian assumption, as the observed signals can be heavy-tailed distributed, or contain outliers.
In this context, a notable line of work addressed the generalization of the aforementioned models to the context of multivariate elliptical distributions, and robust covariance matrix estimation \cite{vogel2011elliptical, liu2012transelliptical, hippert2023learning}.

Despite these recent advances, existing graph learning methods still face computational challenges at high-dimensional scenarios.
The core of the issue lies in the fact that the number of possible edges (thus, the number of variables to estimate) scales to the square of $p$.
Additionally, most iterative algorithms based on GGMs or their variants involve steps that require the inversion (or the singular value decomposition) of $\boldsymbol{\Theta}$.
Thus their complexity that scales in $\mathcal{O}(p^3)$.
Some works mitigated this issue through the assumption of low-rank structures over the covariance matrix $\boldsymbol{\Sigma}$ \cite{hippert2023learning}.
However, these approaches model the covariance matrix, so they still involve tedious inversions steps in a formalism that aims to take control of the sparsity of the graph (i.e. the support of $\boldsymbol{\Theta}=\boldsymbol{\Sigma}^{-1}$).
To address these limitations, we propose an efficient graph learning framework that directly parameterizes $\mathbf{\Theta}$ using a low-rank factorization in order to leverage Riemannian optimization \cite{absil2008optimization, boumal2023introduction}. 
In details, the presented contributions are the following:
\begin{itemize}
    \item 
    We propose a Riemannian optimization-based framework that enables scalable and efficient graph learning by leveraging a low-rank correlation structure on the precision matrix.
    Specifically, we derive Riemannian optimization tools to handle a parameterization of low-rank positive semi-definite matrices that involves the oblique manifold quotiented by rotation matrices.    
    The interest of such an approach is to exploit the conditional correlation structure directly in the graph learning model.
    These tools can be used to perform a Riemannian conjugate gradient algorithm \cite{absil2008optimization, boumal2023introduction} (or more advanced counterparts such as \cite{zhang2016riemannian}) on any generic graph learning objective that involves a low-rank conditionnal correlation.

    \item 
    This framework is then particularized to a low-rank constrained counterpart of the GGMs, i.e., we adapt the GLasso formulation \cite{friedman2008sparse, mazumder2012graphical} in order to account for a low-rank structure over the precision matrix $\boldsymbol{\Theta}$.
    Experiments on synthetic and real data then demonstrate that a very efficient dimension-versus-performance trade-off can be achieved with this approach.

\end{itemize}

This rest of the paper is structured as follows:
Section II presents a general overview of graph learning framework and related works from the graph signal processing literature.
Section III presents the chosen parameterization, and the derivation of the proposed Riemannian optimization framework for large dimensional graph learning.
Section IV validates the proposed approach through experiments on both synthetic data and real-world graph datasets.
Section V concludes the paper with key insights.

The following notation is adopted: italic, lower case boldface, and upper case boldface indicate respectively, scalar, vector, and matrix quantities.
The upperscript $^\top$ denotes the transpose operator.
$\mathbf{I}_p$ is the identity matrix of dimension $p$. 
$\tr(\cdot)$ is the trace operator, and $|\cdot|$ is the determinant one. 
$\{w_i\}_{i=1}^n$ denotes the set of elements $w_i$, $\forall i \in [\![ 1, n ]\!] $.
$\text{diag}(\{a_i\}_{i=1}^n)$ is the $n \times n$ diagonal matrix with diagonal entries $a_n$. $\text{ddiag}(\mathbf{H})$ is the diagonal matrix formed by the main diagonal of the square matrix $\mathbf{H}$.
The set of $p\times p$ symmetric positive definite (resp. semi-definite) matrices is denoted $\mathcal{S}_p^{++}$ (resp. $\mathcal{S}_p^{+}$).
The set of matrices in $\mathcal{S}_p^{+}$ of rank $k$ is denoted  $\mathcal{S}_{p,k}^{+}$.
The oblique manifold of (row-normalized) $p\times k$ matrices is denoted $\mathcal{OB}_{p,k}$, and the orthogonal group of $k\times k$ matrices is denoted $\orthGroup_{\nrank}$.

\section{Overview of graph learning}
\label{sec:overview}
\subsection{Unsupervised graph learning and GLasso}

We consider an undirected graph $\mathcal{G}$ of $p$ nodes, representing $p$ variables denoted by $x_q,~\forall q\in[\![ 1,p ]\!]$.
This graph produces samples that are data vectors, denoted as $\mathbf{x} = [x_1,x_2,\dots,x_p] \in \mathbb{R}^p$.
An edge in $\mathcal{G}$ between the nodes $x_q$ and $x_\ell$ for $q,\ell \in \! [\![1,p]\!] $ exists if ${\rm corr}[x_q x_{\ell} | \mathbf{x}_{ [\![ 1,p ]\!] \backslash \{q,\ell\} } ] \neq 0 $.
Within the graph signal processing framework, a graph learning problem is then related to the estimation of the conditional correlation structure from a sample set $\{\mathbf{x}_i\}_{i=1}^n$, gathered in the data matrix denoted $\mathbf{X}= \left[ \mathbf{x}_1,~\cdots,\mathbf{x}_n \right]\in\mathbb{R}^{p\times n}$.
Gaussian graphical models (GGM) assume that the data is sampled according to $\mathbf{x}\sim \mathcal{N}(\mathbf{0},\boldsymbol{\Sigma})$, with precision matrix $\boldsymbol{\Theta}=\boldsymbol{\Sigma}^{-1}$.
The conditional correlation is then expressed as
\begin{equation}
    {\rm corr}[x_q x_{\ell} | \mathbf{x}_{[\![ 1,p ]\!] \backslash \{q,\ell\} } ]
    =
    - \boldsymbol{\Theta}_{q\ell} / \sqrt{\boldsymbol{\Theta}_{qq}\boldsymbol{\Theta}_{\ell\ell}}.
\label{eq:cond_corr}
\end{equation} 
thus, the conditional correlation structure can be directly linked to the support of the precision matrix $\boldsymbol{\Theta}$.
Graph learning in GGMs is thus usually framed as a structured covariance (or precision) matrix estimation problem.
Indeed, a graph with only few edges per nodes implies that most conditional correlation coefficients are null, and thus a sparse structure on the precision matrix $\boldsymbol{\Theta}$.
In this context, a seminal method was proposed in \cite{friedman2008sparse}, and is referred to as the Graphical Lasso algorithm (GLasso).
This algorithm aims to solve a penalized maximum likelihood estimation for GGM, i.e.,
\begin{equation}  \label{eq:Glasso}
    \underset{\boldsymbol{\Theta}\in \mathcal{S}_p^{++}}{\rm minimize} 
    \quad
    \tr({\mathbf{S}} \boldsymbol{\Theta})
    - \log | \boldsymbol{\Theta} |
    + \lambda || \boldsymbol{\Theta} ||_{1,{\rm off}}, 
\end{equation}
where $\mathbf{S}=\frac{1}{n}\sum_{i=1}^n \mathbf{x}_i\mathbf{x}_i^{\top}$ is the sample covariance matrix, and where the $\ell_1$-norm penalty promotes the sparsity on off-diagonal elements of $\boldsymbol{\Theta}$, whose level is controlled by a regularization parameter $\lambda$.
In details, GLasso performs a block coordinate descent over columns of $\boldsymbol{\Theta}$ while relaxing the constraint $\boldsymbol{\Theta}\in \mathcal{S}_p^{++}$.
However, the problem \eqref{eq:Glasso} (i.e., fitting a GGM with $\ell_1$-norm penalty) is also often referred to as GLasso by association, even if it is addressed with a different optimization algorithm.

\subsection{Beyond GLasso: overview of recent developments}

Many other algorithms were developed to perform graph learning within the aforementioned paradigm.
We propose here an overview by casting the following generic graph learning problem: \vspace{0.6cm}
\begin{equation*} \label{eq:graph_cost}
    \begin{array}{c l}
         \underset{\boldsymbol{\Theta}}{\rm minimize} 
         &  
         \eqnmarkbox[blue]{model}{\mathcal{L}(\boldsymbol{\Theta},\mathbf{X})}
         + 
         \eqnmarkbox[red]{sparsity}{\lambda h(\boldsymbol{\Theta})}
         \vspace{0.2cm} \\
         {\rm subject~to} 
         & 
         \eqnmarkbox[orange]{cov_struct}{\boldsymbol{\Theta} \in \mathcal{S}_{\boldsymbol{\Theta}}}
         ~~{\rm and}~~
         \boldsymbol{\Theta}^{+} = 
         \eqnmarkbox[magenta]{prec_struct}{\boldsymbol{\Sigma} \in \mathcal{S}_{\boldsymbol{\Sigma}}}, \vspace{0.6cm}
    \end{array}
\end{equation*}
\annotate[yshift=0.5em]{above, left}{model}{{\footnotesize Fit graph to data}}
\annotate[yshift=0.5em]{above, right}{sparsity}{{\footnotesize Promote graph sparsity}}
\annotate[yshift=-0.5em]{below, left}{cov_struct}{{\footnotesize Structured precision}}
\annotate[yshift=-0.5em]{below, left}{prec_struct}{{\footnotesize Structured covariance}}

\noindent
where $\mathcal{L}(\boldsymbol{\Theta}, \boldsymbol{X})$ represents the objective function capturing the link between the data matrix $\mathbf{X} \in \mathbb{R}^{p \times n}$ and the graph represented by $\boldsymbol{\Theta}$, $h$ is a regularization term controlled by the parameter $\lambda$, and where $\mathcal{S}_{\boldsymbol{\Theta}}$
(resp. $\mathcal{S}_{\boldsymbol{\Sigma}}$) is a matrix set that models prior knowledge on the structure of $\boldsymbol{\Theta}$ (resp.  $\boldsymbol{\Sigma}$).
In this general set-up, many graph learning algorithms were proposed in order to go beyond the initial assumptions of the regularized GGM model in \eqref{eq:Glasso}.

\subsubsection{Fitting cost $\mathcal{L}$} 
Due to the central limit theorem, the Gaussian distribution is usually taken as likelihood to model stationary signals on graphs \cite[Sec. III.A.]{leus2023graph}.
However, many modern datasets present heavy-tailed distributed data, or contain outliers.
In order to account for this a popular model is brought by multivariate elliptical distributions \cite{cambanis1981theory, delmas2024elliptically}, which allows for linking robust covariance matrix estimation \cite{maronna1976robust, tyler1987distribution} to the problem of graph learning  \cite{wald2019globally, hippert2023learning, de2021graphical, de2024learning}.

\subsubsection{Sparsity promoting penalties $h$}
The $\ell_1$-norm introduced in \cite{friedman2008sparse} (as proxy to the $\ell_0$-norm) remains the most a popular choice for learning sparse graphs.
Still, several studies have explored non-convex penalty functions as alternatives to the $\ell_1$-norm \cite{lam2009sparsistency, shen2012likelihood, benfenati2020proximal}, which have been shown to outperform $\ell_1$-norm when the sparsity parameter $\lambda$ exceeds a certain threshold \cite{ying2020nonconvex}.
Some works also considered penalties that promote specific sparsity patterns \cite{heinavaara2016inconsistency, tarzanagh2018estimation}.

\subsubsection{Matrix structures} \label{subsubsec:overview_structures}
From a statistical standpoint, prior structural knowledge is often expressed as a constraint on the covariance or precision matrix.
Popular low-dimensional linear models include probabilistic principal component analysis \cite{tipping1999probabilistic} and factor models \cite{rubin1982algorithms, robertson2007maximum, khamaru2019computation}, that have been leveraged in graph learning using different approaches \cite{hippert2023learning, chandra2021bayesian, vogels2024bayesian}.
Another popular structure for conditional correlations is the so-called total positivity of order two (MTP2), also referred to as attractive Gaussian graphical models, which imposes the constraint $\mathbf{\Theta}_{ql} \leq 0, \forall q \neq l$ \cite{fallat2017total, Lauritzen2017MaximumLE}. 
Some models alternatively included a normalization constraint to fit the data standardization, i.e., correlation-based structures \cite{phi2024robust, phi2024leveraging}.
The graph signal processing framework also introduced many models of structured covariance and precision matrices \cite{leus2023graph}.
In this scope, the use of the graph Laplacian matrix, related to the notion of smooth graph signal, was first used in order to propose graph learning algorithms \cite{kalofolias2016learn}.
This structure was quickly incorporated into the GLasso formulations, leading to the so-called Laplacian-constrained GMRF \cite{egilmez2017graph, ying2020nonconvex}.
Furthermore, the eigenvalues of the Laplacian graph encode several properties of the graph \cite{mohar1997some}, while its eigenvectors are related to graph filters \cite{isufi2024graph}.
Hence, several works aimed to include prior knowledge on eigenstructure in graph learning \cite{zhao2019optimization, kumar2020unified, buciulea2025polynomial}.

\subsubsection{Optimization} \label{subsubsec:overview_optim}
Given choices of model and structure constraints, the problem expressed in \eqref{eq:graph_cost} is usually hard to solve.
The aforementioned studies relied on different approaches to address it.
For example, Laplacian GRMF were tackled with block-coordinate descent \cite{egilmez2017graph}, majorization-minimization \cite{ying2020nonconvex}, and proximal second-order approaches \cite{medvedovsky2024efficient}.
The framework of Riemannian optimization \cite{absil2008optimization, boumal2023introduction} was used to handle positive definiteness and factor structures in \cite{bouchard2021riemannian, hippert2023learning, phi2024leveraging}.
In many cases, the combination of two structures, or the combination of particular structures with a non-smooth penalty, yields problems in which a single optimization framework can be limiting.
In this case, variable splitting tricks \cite{boyd2011distributed, kovnatsky2016madmm} allow for fragmenting the initial problem into simpler ones.
These have for example been used for graph learning in \cite{zhao2019optimization, kumar2020unified}.

\subsection{Proposed method}

The contributions of this paper relate to the points discussed in \ref{subsubsec:overview_optim} and \ref{subsubsec:overview_structures}: we propose a tractable graph learning framework for large graph structures.
We consider a direct parameterization of the conditional correlation matrix \eqref{eq:cond_corr} as a low-rank one by leveraging a quotient manifold involving the oblique manifold for fixed-rank matrices.
We then leverage the corresponding structure within the Riemannian optimization framework.
Admittedly, this choice is quite restrictive, and is expected to be outperformed by models that have more parameters when large sample support is available.
However our experiments evidence that this structure still holds a favorable trade-off between accuracy and computational complexity. 
In conclusion, this shows that a good performance can still be achieved by a graph learning algorithm that scales well by considerably reducing the dimension of the estimation problem, and efficiently solving it.

\section{Learning graphs leveraging a low-rank correlation structure}
\label{sec:riem_opt}
In this section, we provide a Riemannian optimization-based framework to learn graphs by leveraging a low-rank correlation structure on the precision matrix of the graph.
The interest of such an approach is to exploit the conditional correlation structure of~\eqref{eq:cond_corr} directly in the graph learning model, while reducing its dimension.
After defining the parameter space, which turns out to be a Riemannian manifold, we provide an optimization framework on it.
This framework is then leveraged to develop an original graph learning method.

\subsection{Parameter space}

The choice of parameter space will be motivated by three points:
(\textit{i}) we require a low-rank factorization in order to reduce the number of variables;
(\textit{ii}) we want a direct control over the coefficients in \eqref{eq:cond_corr};
(\textit{iii}) though it will not be studied in this paper, we want to leave the possibility of solving the graph learning problem with block-wise strategies, which, for example, prevents the use of $p$-dimensional eigenvectors (whose the normalization constraint interconnects all variables in the problem).
These motivations lead to the parameterization of the precision matrix $\precision$ as:
\begin{equation}
    \precision = \diag(\powers)\W\W^\top\diag(\powers),
\label{eq:precision_struct}
\end{equation}
where $\diag(\cdot)$ returns the diagonal matrix whose diagonal elements correspond to its argument, $\powers\in\Rplusplus^{\nfeatures}$ (set of vectors with strictly positive elements), and $\W\in\oblique_{\nfeatures,\nrank}$, where the oblique manifold~\cite{absil2006joint} is defined as
\begin{equation}
    \oblique_{\nfeatures,\nrank} =
    \{\W\in\realSpace^{\nfeatures\times\nrank}: \, \ddiag(\W\W^\top)=\eye_{\nfeatures}\}.
\label{eq:oblique}
\end{equation}

\begin{remark}
    With the chosen parametrization~\eqref{eq:precision_struct}, one would expect the rank of the precision matrix to be equal to $\nrank$.
    However, we do not strictly enforce this in the oblique manifold~\eqref{eq:oblique}.
    Thus, within our setting, we have $\rank(\precision)\leq\nrank$.
    However, a proper choice of the objective function will still ensure $\rank(\precision)=\nrank$ (its value will tend to $\infty$ otherwise). 
\end{remark}

Our parametrization~\eqref{eq:precision_struct} admits an invariance: for all $\mathbf{O}\in\orthGroup_{\nrank}$ (the orthogonal group), we have $\W\mathbf{O}(\W\mathbf{O})^\top = \W\W^\top$.
Hence, instead of being on the product manifold $\oblique_{\nfeatures,\nrank}\times\Rplusplus^{\nfeatures}$, we are actually on the product manifold $(\oblique_{\nfeatures,\nrank}/\orthGroup_{\nrank})\times\Rplusplus^{\nfeatures}$, where $\oblique_{\nfeatures,\nrank}/\orthGroup_{\nrank}$ is the quotient manifold of the oblique manifold by the orthogonal group through the equivalence relationship $\W\sim\W\mathbf{O}$, for all $\W\in\oblique_{\nfeatures,\nrank}$ and $\mathbf{O}\in\orthGroup_{\nrank}$.
The manifold that we consider is actually a parametrization of the manifold $\SPSD_{\nfeatures,\nrank}$ of symmetric positive semi-definite matrices of rank $\nrank$.
Geometries for the latter manifold have already been extensively studied in the literature; see, \textit{e.g.},~\cite{bhatia2009positive, bouchard2020riemannian, bouchard2024fisher, mian2024online} for the full rank case ($k=p$), and~\cite{bonnabel2010riemannian,vandereycken2010riemannian,massart2020quotient,bouchard2021riemannian} for the low-rank one.
However, no optimal solution exists from a Riemannian geometry point of view.
Indeed, important geometrical objects remain unknown for some (\textit{e.g.},~\cite{bouchard2021riemannian}), while they cannot guarantee that we stay on the manifold for others (\textit{e.g.},~\cite{massart2020quotient}).
In this work, we thus propose a new parametrization for $\SPSD_{\nfeatures,\nrank}$ that fits well the structure of the graph that we aim to learn.
Indeed, we will have access to the conditional correlation coefficients in \eqref{eq:cond_corr} through the elements of $\W\W^\top$.

The manifold of interest is therefore $\SPSD_{\nfeatures,\nrank} \simeq (\oblique_{\nfeatures,\nrank}/\orthGroup_{\nrank})\times\Rplusplus^{\nfeatures}$.
Notice that the mapping from $\oblique_{\nfeatures,\nrank}\times\Rplusplus^{\nfeatures}$ onto $\SPSD_{\nfeatures,\nrank}$ is
\begin{equation}
    \varphi : (\W,\powers) \mapsto \diag(\powers)\W\W^\top\diag(\powers).
\label{eq:mapping}
\end{equation}
Our parametrization relies on two manifolds: $\oblique_{\nfeatures,\nrank}/\orthGroup_{\nrank}$ and $\Rplusplus^{\nfeatures}$.
For $\Rplusplus^{\nfeatures}$, the Riemannian geometry is very well known.
The manifold $\oblique_{\nfeatures,\nrank}/\orthGroup_{\nrank}$ has been less investigated.
The case $\nfeatures=\nrank$ has been studied in~\cite{thanwerdas2022theoretically}.
Very recently, the Riemannian geometry for $\nrank<\nfeatures$ has been treated in~\cite{chen2025quotient}.
In this article, our angle is different because we are more interested in the optimization aspects than in the geometrical modeling.
For example, we will propose efficient retractions over studying geodesics.

\subsection{Riemannian optimization on $(\oblique_{\nfeatures,\nrank}/\orthGroup_{\nrank})\times\Rplusplus^{\nfeatures}$}

When dealing with a product manifold, one can deal with each manifold separately and then gather all objects together.
Given a manifold, to be able to perform Riemannian optimization, we need the following objects:
\textit{(i)} representations for the tangent spaces;
\textit{(ii)} a Riemannian metric, which then yields the definition of Riemannian gradient;
\textit{(iii)} the orthogonal projection from the ambient space (Euclidean space the manifold lives in) onto tangent spaces, which is very useful to obtain the Riemannian gradient from the Euclidean one;
\textit{(iv)} a retraction, which maps tangent vectors onto the manifold, hence allowing to obtain new iterates from descent directions.
These tools will be detailed for both manifolds $\oblique_{\nfeatures,\nrank}/\orthGroup_{\nrank}$ and $\Rplusplus^{\nfeatures}$ in the following subsections.

\subsubsection{$\oblique_{\nfeatures,\nrank}/\orthGroup_{\nrank}$}
To handle the geometry of a quotient manifold, one exploits representations in the parent manifold, which, in our case, is $\oblique_{\nfeatures,\nrank}$.
Thus, characterizing the geometry of the quotient manifold is also crucial.
Full reviews on manipulating quotient manifolds can be found in~\cite{absil2008optimization,boumal2023introduction}.

The first thing that one needs is the tangent space $\obliqueTangent{\W}_{\nfeatures,\nrank}$ at $\W\in\oblique_{\nfeatures,\nrank}$.
It is~\cite{absil2006joint}
\begin{equation}
    \obliqueTangent{\W}_{\nfeatures,\nrank} = \{ \Wtangent\in\realSpace^{\nfeatures\times\nrank} : \, \ddiag(\Wtangent\W^\top) = \zero \}.
\label{eq:ob_metric}
\end{equation}
One then needs a Riemannian metric on $\oblique_{\nfeatures,\nrank}$.
In our case, we can simply employ the Euclidean metric
\begin{equation}
    \metric{\W}{\Wtangent}{\WtangentBis}^{\oblique} = \tr(\Wtangent^\top\WtangentBis).
\end{equation}
Indeed, as needed, it is invariant along equivalence classes, \textit{i.e.}, for all $\mathbf{O}\in\orthGroup_{\nrank}$,
\begin{equation}
    \metric{\W\mathbf{O}}{\Wtangent\mathbf{O}}{\WtangentBis\mathbf{O}}^{\oblique} = \metric{\W}{\Wtangent}{\WtangentBis}.
\end{equation}
From there, a representation of the tangent space of $\oblique_{\nfeatures,\nrank}/\orthGroup_{\nrank}$ can be obtained in the equivalence class of $\W$, $\pi(\W)=\{\W\mathbf{O}: \mathbf{O}\in\orthGroup_{\nrank}\}$.
In fact, it is represented by a subspace of $\obliqueTangent{\W}_{\nfeatures,\nrank}$: the so-called horizontal space $\horizontal{\W}$.
To obtain it, one first needs the vertical space $\vertical{\W}$, which is the tangent space $T_{\W}\pi(\W)$ of the equivalence class.
In our case, it is
\begin{equation}
    \vertical{W} = \{\W\boldsymbol{\Omega}: \, \boldsymbol{\Omega}\in\mathcal{A}_{\nrank}\},
\label{eq:ob_vertical}
\end{equation}
where $\mathcal{A}_{\nrank}$ is the space of skew-symmetric matrices.
Notice that $\vertical{\W}\subset\obliqueTangent{\W}$ since the congruence of a skew-symmetric matrix yields a skew-symmetric matrix.
The formula for the vertical space $\vertical{W}$ directly arises from the fact that $T_{\mathbf{O}}\orthGroup_{\nrank}=\{\mathbf{O}\boldsymbol{\Omega}: \, \boldsymbol{\Omega}\in\mathcal{A}_{\nrank}\}$; see, \textit{e.g.},~\cite{edelman1998geometry,absil2008optimization}.
The horizontal space $\horizontal{\W}$ is then defined as the orthogonal complement to $\vertical{\W}$ in $\obliqueTangent{\W}_{\nfeatures,\nrank}$ according to the Riemannian metric.
To ensure that $\Wtangent\in\obliqueTangent{\W}$ is orthogonal to the vertical space~\eqref{eq:ob_vertical} according to~\eqref{eq:ob_metric}, it is necessary and sufficient for $\W^\top\Wtangent$ to be symmetric.
Therefore, the horizontal space $\horizontal{\W}$ is
\begin{multline}
    \horizontal{\W} = \{ \Wtangent\in\realSpace^{\nfeatures\times\nrank} : \, \ddiag(\Wtangent\W^\top) = \zero, \\ \Wtangent^\top\W = \W^\top\Wtangent \}/.
\end{multline}

When it comes to optimization, an object that is particularly useful is the orthogonal projection on the tangent space according to the chosen metric.
Indeed, it allows to get the Riemannian gradient of an objective function from the Euclidean one.
From~\cite{absil2006joint}, the projection from $\realSpace^{\nfeatures\times\nrank}$ onto $\obliqueTangent{\W}$ is
\begin{equation}
    P^{\oblique}_{\W}(\mathbf{Z}) = \mathbf{Z} - \ddiag(\mathbf{Z}\mathbf{W}^\top)\mathbf{W}.
\label{eq:ob_proj}
\end{equation}
As explained in the next paragraph, the projection~\eqref{eq:ob_proj} is enough to obtain the Riemannian gradient thanks to an invariance property of the objective function.
However, if one wants to employ a more sophisticated method than the Riemannian gradient descent, such as the conjugate gradient or the BFGS algorithm~\cite{absil2008optimization,boumal2023introduction}, one also needs the orthogonal projection on the horizontal space $\horizontal{\W}$.
As noted in~\cite{chen2025quotient}, the projection derived in~\cite{massart2020quotient} on a similar quotient (the difference is that it does not include the constraint $\ddiag(\W\W^\top)=\eye_{\nfeatures}$) remains valid in our case.
To obtain it, consider the decomposition of $\Wtangent\in\obliqueTangent{\W}$:
\begin{equation}
    \Wtangent = \W\boldsymbol{\Omega} + \mathbf{Y},
\end{equation}
where $P^{\vertical{}}_{\W}(\Wtangent) = \W\boldsymbol{\Omega}$ is the orthogonal projection on the vertical space and $P^{\horizontal{}}_{\W}(\Wtangent) = \mathbf{Y}$ is the projection on the horizontal space.
Recalling $\W^\top\mathbf{Y} = \mathbf{Y}^\top\W$, one obtains
\begin{equation}
    \W^\top\Wtangent - \Wtangent^\top\W = \W^\top\W\boldsymbol{\Omega} + \boldsymbol{\Omega}\W^\top\W.
\end{equation}
As explained in~\cite{massart2020quotient}, this is a Sylvester equation that admits a unique solution $\boldsymbol{\Omega}\in\mathcal{A}_{\nrank}$, which can be efficiently computed.
From there, the orthogonal projection on $\horizontal{\W}$ is
\begin{equation}
    P^{\horizontal{}}_{\W}(\Wtangent) = \Wtangent - \W\boldsymbol{\Omega}.
\end{equation}

These objects are enough to define the Riemannian gradient of an objective function $\bar{f}:\oblique_{\nfeatures,\nrank}/\orthGroup_{\nrank}\to\realSpace$ and its representation on $\oblique_{\nfeatures,\nrank}$ $f=\bar{f}\circ\pi:\oblique_{\nfeatures,\nrank}\to\realSpace$.
The Riemannian gradient of an objective function $f$ at $\W$ is defined as the unique tangent vector $\grad f(\W)$ such that, for all $\Wtangent\in\obliqueTangent{\W}_{\nfeatures,\nrank}$, $\metric{\W}{\grad f(\W)}{\Wtangent} = \D f(\W)[\Wtangent]$, where $\D f(\W)[\Wtangent]$ is the directional derivative of $f$ at $\W$ in direction $\Wtangent$~\cite{absil2008optimization,boumal2023introduction}.
It is possible to obtain the Riemannian gradient $\grad^{\oblique} f(\W)$ from the Euclidean gradient $\grad^{\mathcal{E}} f(\W)$ through the projection
\begin{equation}
    \grad^{\oblique} f(\W) = P_{\W}(\grad^{\mathcal{E}} f(\W)).
\end{equation}
The Riemannian gradient of $\bar{f}$ is actually simply represented by the Riemannian gradient of $g$.
Interestingly, at $\W\in\oblique_{\nfeatures,\nrank}$, the Riemannian gradient $\grad^{\oblique} f(\W)$ of $f$ already belongs to $\horizontal{\W}$, so it is not necessary to project it.
This arises from the fact that, to correspond to a proper objective function $\bar{f}$ on the quotient, $f$ must be invariant along equivalence classes, \textit{i.e.}, for all $\mathbf{O}\in\orthGroup$, $f(\W)=f(\W\mathbf{O})$ (hence its Riemannian gradient does not have a component on the vertical space).

To be able to apply a Riemannian gradient descent algorithm on $\oblique_{\nfeatures,\nrank}/\orthGroup_{\nrank}$, it remains to define a retraction, which maps tangent vectors back onto the manifold.
As for the rest, a representation on $\oblique_{\nfeatures,\nrank}$ can be leveraged.
A retraction $R^{\oblique}$ on $\oblique_{\nfeatures,\nrank}$ yields a proper retraction on $\oblique_{\nfeatures,\nrank}/\orthGroup_{\nrank}$ if it is invariant along equivalence classes, \textit{i.e.}, if, for all $\W\in\oblique_{\nfeatures,\nrank}$ and $\mathbf{O}\in\orthGroup_{\nrank}$, $\pi(R^{\oblique}_{\W\mathbf{O}}(\Wtangent\mathbf{O})) = \pi(R^{\oblique}_{\W}(\Wtangent))$: there exists $\mathbf{\bar{O}}\in\oblique_{\nfeatures,\nrank}$ such that $R^{\oblique}_{\W\mathbf{O}}(\Wtangent\mathbf{O})=R^{\oblique}_{\W}(\Wtangent)\mathbf{\bar{O}}$.
In this paper, we consider the projection-based retraction framework proposed in~\cite{absil2012projection} to define an adequate retraction on $\oblique_{\nfeatures,\nrank}$.
It is
\begin{equation}
    R^{\oblique}_{\W}(\Wtangent) = \mathcal{P}(\W+\Wtangent),
\label{eq:ob_retr}
\end{equation}
where the projection $\mathcal{P}:\realSpace^{\nfeatures\times\nrank}\to\oblique_{\nfeatures,\nrank}$ is
\begin{equation}
    \mathcal{P}(\mathbf{Z}) = \ddiag(\mathbf{Z}\mathbf{Z}^\top)^{-1/2}\mathbf{Z}.
\end{equation}
It is readily checked that $\mathcal{P}(\mathbf{Z}\mathbf{O})=\mathcal{P}(\mathbf{Z})\mathbf{O}$.
Hence, \eqref{eq:ob_retr} produces a proper retraction on the quotient.

The geometrical objects above are enough to employ the Riemannian gradient descent algorithm on $\oblique_{\nfeatures,\nrank}/\orthGroup_{\nrank}$.
However, if one wants to use more sophisticated optimization methods such as Riemannian conjugate gradient or BFGS, a vector transport operator is also needed~\cite{absil2008optimization,boumal2023introduction}.
Such an object allows to transport a tangent vector from the tangent space of a point onto the tangent space of another point.
This is very useful when it comes to exploiting previous descent directions in the optimization process.
As before, we rely on a representation on $\oblique_{\nfeatures,\nrank}$.
From~\cite{absil2008optimization}, the simplest (generic) solution to transport $\Wtangent$ from $\obliqueTangent{\W}$ to $\obliqueTangent{\bar{\W}}$ is
\begin{equation}
    \mathcal{T}^{\oblique}_{\W\to\bar{\W}}(\Wtangent) = P^{\horizontal{}}_{\bar{\W}}(\Wtangent).
\end{equation}

\subsubsection{$\Rplusplus^{\nfeatures}$}
The geometry of $\Rplusplus^{\nfeatures}$ is more straightforward and can also be found for instance in~\cite{bouchard2020riemannian,mian2024online}.
This manifold is open in the vector space $\realSpace^{\nfeatures}$.
Thus, the tangent space $T_{\powers}\Rplusplus^{\nfeatures}$ at any $\powers\in\Rplusplus^{\nfeatures}$ can be identified to $\realSpace^{\nfeatures}$.
The most natural Riemannian metric to adequately take into account the geometrical structure of $\Rplusplus^{\nfeatures}$ is, for $\powers\in\Rplusplus^{\nfeatures}$, $\powersTangent$ and $\powersTangentBis\in\realSpace^{\nfeatures}$
\begin{equation}
    \metric{\powers}{\powersTangent}{\powersTangentBis}^{\Rplusplus} = \powersTangent^\top (\powers^{\odot -2} \odot \powersTangentBis),
\end{equation}
where $\odot$ denotes the Hadamard product and $\cdot^{\odot \cdot}$ the element-wise power function.
Notice that this metric is the counterpart of the affine-invariant metric on the manifold of symmetric positive definite matrices~\cite{bhatia2009positive,bouchard2024fisher}.

Since the tangent space at $\powers\in\Rplusplus^{\nfeatures}$ is the whole vector space $\realSpace^{\nfeatures}$, no projection is needed in this case.
However, in this case, because the chosen metric is different from the Euclidean one, we still need to transform the Euclidean gradient in order to get the Riemannian one.
Given an objective function $f:\Rplusplus^{\nfeatures}\to\realSpace$, the Riemannian gradient is obtained from the Euclidean one with the formula
\begin{equation}
    \grad^{\Rplusplus} f(\powers) = \powers^{\odot 2} \odot \grad^{\mathcal{E}} f(\powers).
\end{equation}
Concerning the retraction, the best choice%
\footnote{
    To ensure the geometrical constraint while guaranteeing numerical efficiency and stability.
}
is to take the second-order approximation of the Riemannian exponential map, which is
\begin{equation}
    R^{\Rplusplus}_{\powers}(\powersTangent) = \powers + \powersTangent + \frac12 \powers^{\odot -1} \odot \powersTangent^{\odot 2}.
\end{equation}
Finally, for the vector transport, we chose the one resulting from the parallel transport, which is, in this case,
\begin{equation}
    \mathcal{T}^{\Rplusplus}_{\powers\to\bar{\powers}}(\powersTangent) = \bar{\powers} \odot \powers^{\odot -1} \odot \powersTangent.
\end{equation}

\subsubsection{$(\oblique_{\nfeatures,\nrank}/\orthGroup_{\nrank})\times\Rplusplus^{\nfeatures}$}
Dealing with the product manifold $(\oblique_{\nfeatures,\nrank}/\orthGroup_{\nrank})\times\Rplusplus^{\nfeatures}$ is straightforward once geometrical objects have been defined on each component, \textit{i.e.}, $(\oblique_{\nfeatures,\nrank}/\orthGroup_{\nrank})$ and $\Rplusplus^{\nfeatures}$.
The tangent space at $(\W,\powers)$ is represented by the product $\horizontal{\W}\times\realSpace^{\nfeatures}$.
For the Riemannian metric, one can simply take the sum of both metrics, \textit{i.e.},
\begin{multline}
    \metric{(\W,\powers)}{(\Wtangent,\powersTangent)}{(\WtangentBis,\powersTangentBis)}
    =
    \metric{\W}{\Wtangent}{\WtangentBis}^{\oblique}
    \\
    + \metric{\powers}{\powersTangent}{\powersTangentBis}^{\Rplusplus}.
\end{multline}
All other objects are obtained by encapsulating the objects of each manifold in tuples.
For instance, the projection map is
\begin{equation}
    P_{(\W,\powers)}(\mathbf{Z}_{\W},\mathbf{z}_{\powers}) = (P^{\horizontal{}}_{\W}(P^{\oblique}_{\W}(\mathbf{Z}_{\W})),\mathbf{z}_{\powers}).
\end{equation}
Given an objective function $\bar{f}:(\oblique_{\nfeatures,\nrank}/\orthGroup_{\nrank})\times\Rplusplus^{\nfeatures}\to\realSpace$ with corresponding objective function $f:\oblique_{\nfeatures,\nrank}\times\Rplusplus^{\nfeatures}\to\realSpace$, the Riemannian gradient can be obtained from the Euclidean one $\grad^{\mathcal{E}} f(\W,\powers) = (\grad^{\mathcal{E}} f(\W), \grad^{\mathcal{E}} f(\powers))$ through
\begin{multline}
    \grad f(\W,\powers) = (\grad^{\oblique} f(\W), \grad^{\Rplusplus} f(\powers))
    \\
    = (P^{\oblique}_{\W}(\grad^{\mathcal{E}} f(\W)), \powers^{\odot 2} \odot \grad^{\mathcal{E}} f(\powers)).
\label{eq:grad_quotient}
\end{multline}
Finally, the retraction and vector transport are 
\begin{equation}
    \begin{array}{l}
         R_{(\W,\powers)}(\Wtangent,\powersTangent)
         = (R^{\oblique}_{\W}(\Wtangent),R^{\Rplusplus}_{\powers}(\powersTangent)),
         \\[5pt]
         \mathcal{T}_{(\W,\powers)\to(\bar{\W},\bar{\powers})}(\Wtangent,\powersTangent)
         = (\mathcal{T}_{\W\to\bar{\W}}(\Wtangent),\mathcal{T}_{\powers\to\bar{\powers}}(\powersTangent)).
    \end{array}
\end{equation}
From there, we have all necessary objects to perform Riemannian optimization on $(\oblique_{\nfeatures,\nrank}/\orthGroup_{\nrank})\times\Rplusplus^{\nfeatures}$ with Riemannian gradient descent, conjugate gradient, BFGS, \textit{etc}.
For example, the Riemannian gradient descent algorithm is
\begin{equation}
\label{eq:alpha}
    \begin{array}{r}
        (\W_{(t+1)},\powers_{(t+1)}) =
        R_{(\W_{(t)},\powers_{(t)})}(-\alpha_t\grad^{\oblique} f(\W_{(t)}),
        \\
        -\alpha_t\grad^{\Rplusplus} f(\powers_{(t)})),
    \end{array}
\end{equation}
where $\alpha_t$ is the step-size, which can be computed through a line-search~\cite{absil2008optimization,boumal2023introduction}.

\subsection{Graph learning on $(\oblique_{\nfeatures,\nrank}/\orthGroup_{\nrank})\times\Rplusplus^{\nfeatures}$}

Now that we have a proper Riemannian optimization framework on $(\oblique_{\nfeatures,\nrank}/\orthGroup_{\nrank})\times\Rplusplus^{\nfeatures}$, we can exploit it in order to learn a graph using the parametrization~\eqref{eq:precision_struct}.
In this context, we have an objective function of the form~\eqref{eq:graph_cost} on $\SPSD_{\nfeatures,\nrank}$.
Let $g:\SPSD_{\nfeatures,\nrank}\to\realSpace$ be such an objective function.
The objective function on $(\oblique_{\nfeatures,\nrank}/\orthGroup_{\nrank})\times\Rplusplus^{\nfeatures}$ is obtained by leveraging the mapping $\varphi$ defined in~\eqref{eq:mapping}.
One gets the objective function $f:\oblique_{\nfeatures,\nrank}\times\Rplusplus^{\nfeatures}\to\realSpace$ and the corresponding function $\bar{f}=f\circ\pi:(\oblique_{\nfeatures,\nrank}/\orthGroup_{\nrank})\times\Rplusplus^{\nfeatures}\to\realSpace$ through $f=g\circ\varphi$.
It appears pretty convenient to obtain a formula that links the Euclidean gradient of $f$ (in $\realSpace^{\nfeatures\times\nrank}\times\realSpace^{\nfeatures}$) to the Euclidean gradient of $g$ (in $\realSpace^{\nfeatures\times\nfeatures}$).
This is obtained by noticing that $\D f(\W,\powers)[\Wtangent,\powersTangent] = \D g(\varphi(\W,\powers))[\D\varphi(\W,\powers)[\Wtangent,\powersTangent]]$ and solving the equation
\begin{multline}
    \tr(\grad^{\mathcal{E}} g(\varphi(\W,\powers)) \D\varphi(\W,\powers)[\Wtangent,\powersTangent])
    =
    \\
    \tr(\grad^{\mathcal{E}} f(\W)^\top \Wtangent)
    +
    \grad^{\mathcal{E}} f(\powers)^\top \powersTangent.
\end{multline}
From
\begin{multline}
    \D\varphi(\W,\powers)[\Wtangent,\powersTangent] =
    \\
    \diag(\powersTangent)\W\W^\top\diag(\powers)
    + \diag(\powers)\W\W^\top\diag(\powersTangent)
    \\
    + \diag(\powers)(\W\Wtangent^\top+\Wtangent\W^\top)\diag(\powers),
\end{multline}
one obtains
\begin{equation}
    \begin{array}{l}
        \grad^{\mathcal{E}} f(\W) = 2 \diag(\powers) \grad^{\mathcal{E}} g(\varphi(\W,\powers)) \diag(\powers) \W
        \\
        \grad^{\mathcal{E}} f(\powers) = 2\diag(\W\W^\top\diag(\powers)\grad^{\mathcal{E}} g(\varphi(\W,\powers))).
    \end{array}
\label{eq:grad_mapping}
\end{equation}

This means that to be able to perform the optimization of $\bar{f}$ on $(\oblique_{\nfeatures,\nrank}/\orthGroup_{\nrank})\times\Rplusplus^{\nfeatures}$, one only needs to compute the Euclidean gradient of $g:\SPSD_{\nfeatures,\nrank}\to\realSpace$.
In this paper, given some samples $\dataMat=[\data_i]_{i=1}^{\nsamples}\in\realSpace^{\nfeatures\times\nsamples}$ we consider the objective function
\begin{equation}
    g(\precision) = \mathcal{L}(\precision,\dataMat) + \lambda h(\precision),
\end{equation}
where $\lambda>0$; $\mathcal{L}$ is closely related to the negative log-likelihood of the multivariate Gaussian distribution with covariance matrix $\precision^{-1}$; and $h$ is a sparsity promoting penalty, which is a smooth surrogate of the $\ell_1$ norm, that was validated in~\cite{hippert2023learning}, \textit{i.e.},
\begin{equation}
    h(\precision) = \sum_{q\neq\ell} \phi([\precision]_{q\ell}),
\end{equation}
where
\begin{equation}
    \phi(t) = \varepsilon \log(\cosh(t/\varepsilon)),
\end{equation}
and $\varepsilon>0$.
The $\ell_1$ norm is obtained by taking $\lim_{\varepsilon\to0}\phi(t)$).
Concerning $\mathcal{L}$, the difference with the negative log-likelihood of the multivariate Gaussian distribution is that we take a truncated version of the determinant, \textit{i.e.},
\begin{equation}
    \mathcal{L}(\precision,\dataMat) = \frac12\tr(\precision\SCM) - \frac12\log\det{}_{\nrank}(\precision),
    \label{eq:LRCC_problem}
\end{equation}
with $\SCM=\frac{1}{\nsamples}\dataMat\dataMat^\top$ and $\det{}_{\nrank}(\precision)=\prod_{q=1}^{\nrank}\lambda_q$, where $\{\lambda_q\}_{q=1}^{\nrank}$ are the $\nrank$ non-zero eigenvalues of $\precision$.
Finally, the Euclidean gradient of $g$ at $\precision$ is
\begin{equation}
    \grad^{\mathcal{E}} g(\precision) = \SCM - \precision^{\dagger} + \lambda\grad^{\mathcal{E}} h(\precision),
\label{eq:grad_g}
\end{equation}
where $\cdot^{\dagger}$ denotes the Moore-Penrose pseudo-inverse, and $\grad^{\mathcal{E}} h(\precision)\in\realSpace^{\nfeatures\times\nfeatures}$, whose elements are
\begin{equation}
    [\grad^{\mathcal{E}} h(\precision)]_{q\ell}=\phi'(\precision_{q\ell})=\tanh(\precision_{q\ell}/\varepsilon).
\end{equation}
Our proposed algorithm is summarized in Algorithm~\ref{algo:corr_graph}, and the corresponding graph learning method will be referred to as LRCC (for low-rank conditionnal correlation).

\begin{algorithm}[t]
    \caption{LRCC graph learning}
    \label{algo:corr_graph}
    \begin{algorithmic}[1]
        \STATEx {\bfseries Input:} data samples $\dataMat=[\data_i]\in\realSpace^{\nfeatures\times\nsamples}$, initial guess $(\W_{(0)},\powers_{(0)})\in(\oblique_{\nfeatures,\nrank}/\orthGroup_{\nrank})\times\Rplusplus^{\nfeatures}$
        \STATE Compute sample covariance matrix $\SCM=\frac{1}{\nsamples}\dataMat\dataMat^\top$
        \STATE Set $t=0$
        \REPEAT
            \STATE Compute $\precision_{(t)}=\varphi(\W_{(t)},\powers_{(t)})$ with~\eqref{eq:mapping}
            \STATE Compute $\grad^{\mathcal{E}} g(\precision_{(t)})$ with~\eqref{eq:grad_g}
            \STATE Deduce $(\grad^{\mathcal{E}} f(\W_{(t)}),\grad^{\mathcal{E}} f(\powers_{(t)}))$ with~\eqref{eq:grad_mapping}
            \STATE Compute $\grad f(\W_{(t)},\powers_{(t)})$ from~\eqref{eq:grad_quotient}
            \STATE Deduce a descent direction $(\boldsymbol{\xi}_{\W_{(t)}},\boldsymbol{\xi}_{\powers_{(t)}})$ of $f$ from $\grad f(\W_{(t)},\powers_{(t)})$ and possibly previous descent directions (conjugate gradient, BFGS, \textit{etc.})
            \STATE $(\W_{(t+1)},\powers_{(t+1)}) = R_{(\W_{(t)},\powers_{(t)})}(\alpha_t\boldsymbol{\xi}_{\W_{(t)}},\alpha_t\boldsymbol{\xi}_{\powers_{(t)}})$, where $\alpha_t$ is computed through a line-search
            \STATE Set $t=t+1$
        \UNTIL{
            convergence
        }
        \STATEx {\bfseries Return:} $(\W_{(t)},\powers_{(t)})$
    \end{algorithmic}
\end{algorithm}

\subsection{Complexity Analysis}
\label{sec:Time_Complexity}

The following analysis will be conducted in terms of cost per iteration.
First we remark that all algorithms based on the SCM $\mathbf{S}$ and the Gaussian log-likelihood involve operations that scale in $\mathcal{O}(n p^2)$.
These are related to matrix multiplications that can be parallelized, which do not represent a computational bottleneck, nor a differentiating point between methods.
The major differences appear when considering the most complex operation required to to compute one iterate.
In our case, this relates to the retraction and computation of the parallel transport, which scales in $\mathcal{O}(pk^2+k^3)$.
In comparison, GLasso \cite{friedman2008sparse} and Laplacian learning methods \cite{kumar2020unified} report a complexity that scales in $\mathcal{O}(p^3)$, which is due to the required inversion (or the singular value decomposition) of $p\times p$ matrices related to $\boldsymbol{\Theta}$.
Factor models \cite{hippert2023learning} also scale in $\mathcal{O}(pk^2+k^3)$ thanks to a low-rank factorization of the covariance matrix.
However, there are two major differences between the two approaches:
(\textit{i}) operations in factor models from \cite{hippert2023learning} involve parameters of the singular value decomposition, which does not trivially allows for parallelism;
(\textit{ii}) models in \cite{hippert2023learning} rely on a parameterization of $\boldsymbol{\Theta}^{-1}$.
Hence, each step requires tedious matrix inversions in order to evaluate both the objective and gradient expressions (that depends both on the parameter and its inverse).
This implies a non-negligible overhead, especially when computing the line-search.
Consequently, although both GGFM and our method share the same theoretical bottleneck complexity of $\mathcal{O}(pk^2 + k^3)$, our proposed approach appears as the most efficient in terms of computational burden.

\section{Experiments}
\label{sec:exp}
In this section, we evaluate the performance of our proposed method (LRCC) to other state-of-the-art methods on both synthetic and real-world datasets. 
Section \ref{subsec:synth_ROC} presents experiments on synthetic data, in which we study the sensitivity of the proposed method to its parameters.
Section \ref{subsec:synth_ROC} presents experiments related to the link prediction task on a sensor network dataset.
Section \ref{subsec:amimals_data} assesses the effectiveness of node clustering visually on a semantic dataset.

\subsection{Link prediction and scalability on synthetic data}
\label{subsec:synth_ROC}

This section evaluates the performance of LRCC and its sensitivity to the choices of the rank $k$, and the regularization parameter $\lambda$.
In particular, we will show that leveraging low-rank factorizations can significantly reduce the dimension of a graph learning model without sacrificing overall performance.\\

\noindent
\textbf{Setup}:
With a predefined data dimension $p$, the graph topology is sampled from the Barabási-Albert (BA) following from \cite{kumar2020unified, hippert2023learning}: edge creation probability is set to $0.1$, and edge weights are drawn from \( U(2,5) \).
This sampling process generates a weighted adjacency matrix $\mathbf{A} \in \mathbb{R}^{p \times p}$, where $ [\mathbf{A}]_{ij} \neq 0 $ if an edge exists between two variables $( i \neq j $), and $[\mathbf{A}]_{ij} = 0$ otherwise. 
For each Monte-Carlo sampling, we generate a graph adjacency matrix $ \mathbf{A} $ from the aforementioned model. The corresponding  graph Laplacian matrix is $\mathbf{L} = \mathbf{D} - \mathbf{A}$, where $ \mathbf{D} $ is the degree matrix, defined as  
$[\mathbf{D}]_{ii} = \sum_{i \neq j} [\mathbf{A}]_{ij}, \quad [\mathbf{D}]_{ij} = 0 \quad \text{for } i \neq j$.
A corresponding covariance matrix $ \mathbf{\Sigma}$ is constructed form the inverse of the precision matrix $\mathbf{\Theta} = \mathbf{L} + \kappa \mathbf{I}_p$, where $ \kappa = 10^{-1} $ ensures non-singularity.
Next, we draw the data $ \{\mathbf{x}_i\}_{i=1}^n $ from a Gaussian graphical model (GGM) $\mathbf{x} \sim \mathcal{N}(\mathbf{0}, \mathbf{\Sigma})$.\\

\noindent
\textbf{Baselines}:
The comparison includes various methods that impose additional structures to the original formulation of GLasso \eqref{eq:Glasso}.
The algorithm GGM solves \eqref{eq:Glasso} with a Riemannian optimization algorithm proposed in \cite{hippert2023learning}, and serves as an equivalent to GLasso.
The algorithm NGL \cite{ying2020nonconvex}, which performs regularized MLE of Laplacian contrained GGM, \textit{i.e.}, GGM with the additional constraint that $\boldsymbol{\Theta}$ is a Laplacian matrix (and a modified sparsity promoting penalty).
The algorithm GGFM, tackles \eqref{eq:Glasso} under the constraint of factor model of rank $k$ on the covariance matrix.
The proposed algorithm LRCC, which solves \ref{eq:LRCC_problem}, \textit{i.e.}, adapts GGM to directly impose a low-rank factorization of $\boldsymbol{\Theta}$.
All methods depend on a regularization parameter $\lambda$.
In order to provide a fair comparison, this parameter is selected for each method based on the highest AUC obtained through a grid search across $50$ Monte Carlo trials.  
For each setup and method, this optimal $ \lambda $ value is determined independently and used for the final reported results.\\

\noindent
\textbf{Evaluation metrics}:
Each method produces an estimate of the precision matrix, that is normalized.  
To determine the corresponding graph topology, a detection threshold within the range $[0,1]$ is applied to the estimated values.  
For each method, we analyze the ROC curves, which illustrate the trade-off between the false positive rate (incorrectly identified edges) and the true positive rate (correctly detected edges) as the threshold varies.  
In addition, the AUC serves as a comprehensive metric for method comparison.\\

\begin{figure*}[t!]
\label{fig: ROC}
\clearpage
    \centering
    \begin{minipage}{0.326\textwidth}
        \centering
        \includegraphics[width=\textwidth]{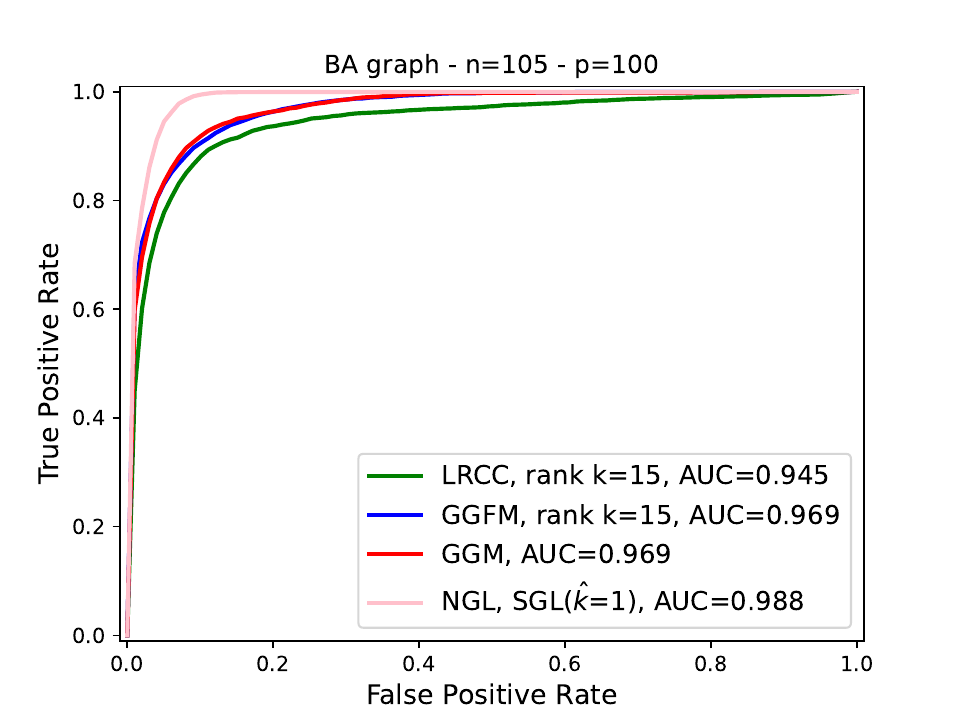}
    \end{minipage}
    \hfill
    \begin{minipage}{0.326\textwidth}
        \centering
        \includegraphics[width=\textwidth]{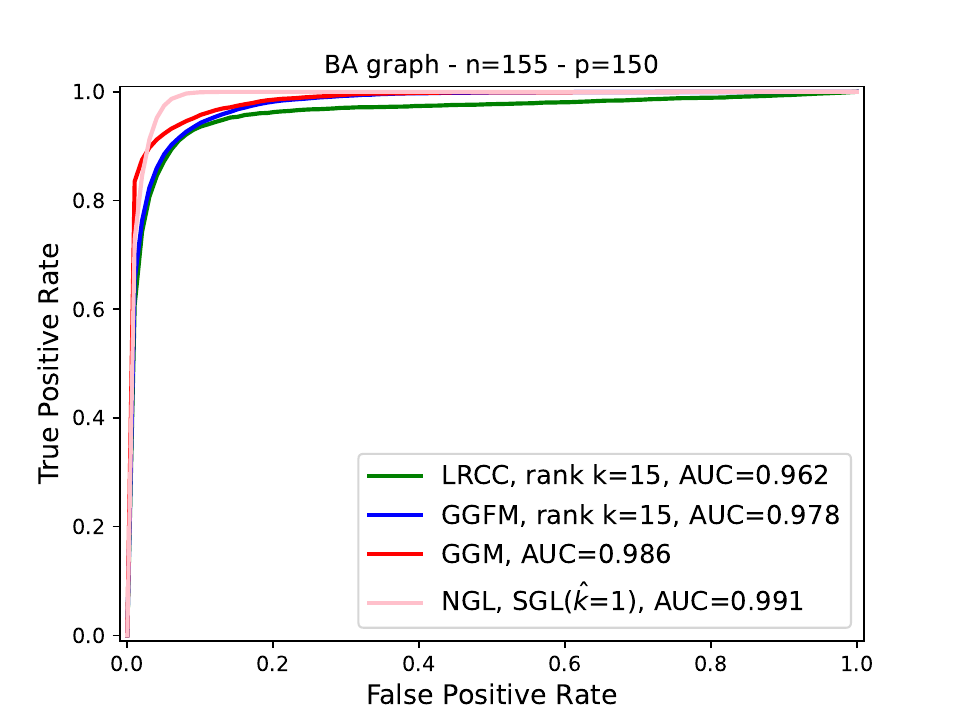}
    \end{minipage}
    \hfill
    \begin{minipage}{0.326\textwidth}
        \centering
        \includegraphics[width=\textwidth]{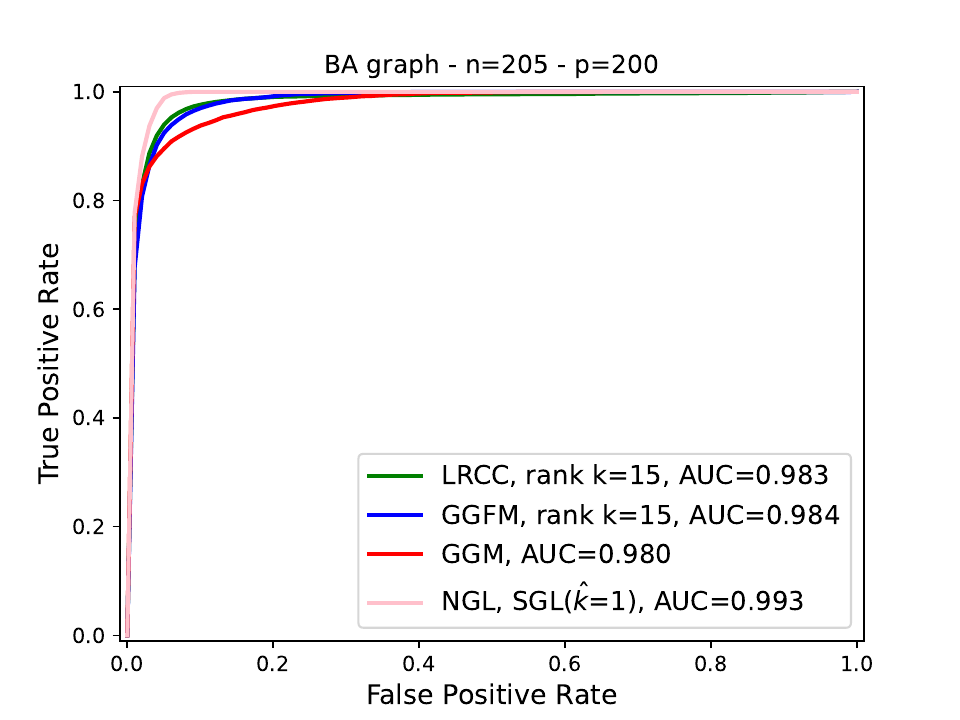}
    \end{minipage}

    \caption{The comparison of ROC/AUC in low-sample support setting $n=p+5$ for Barabási-Albert (BA) graph. From left to right, the settings are $p=100,~p=150,~p=200$, respectively, with the tested rank $k=0.1p$.} 
    \label{fig:roc}
\end{figure*}

\begin{figure}[t!]
\label{fig: auc }
\clearpage
    \centering
    \includegraphics[width=0.5\textwidth]{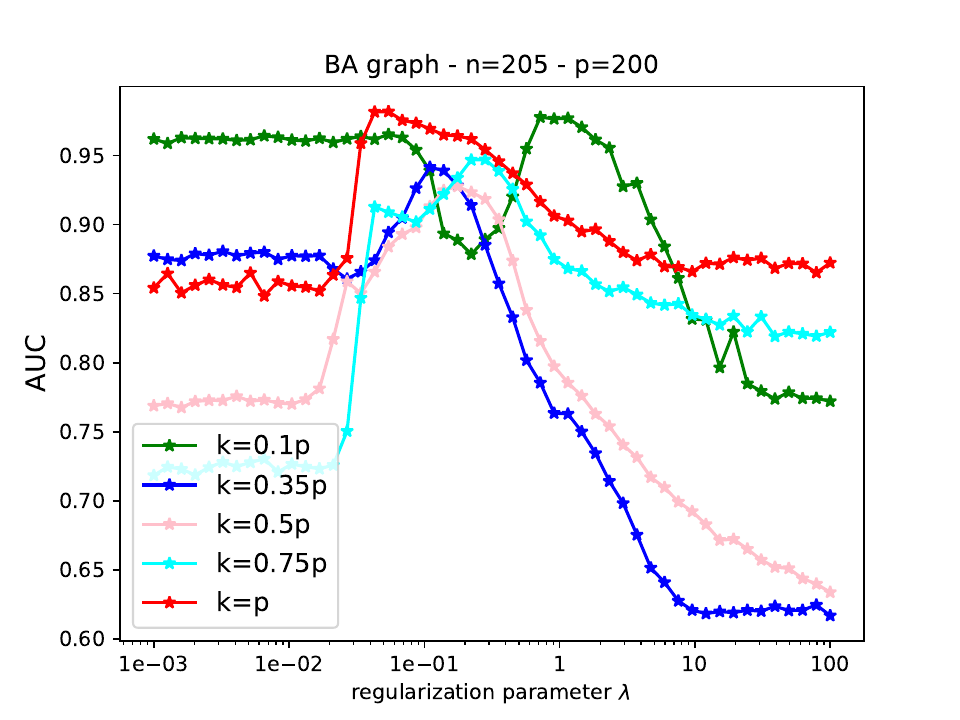}
    \vspace{-0.5cm}
    \caption{The sensitivity of AUC score w.r.t $\lambda \in [10^{-3}, 10^2]$ of LRCC for various ranks $k$ with dimension $p=200$.} 
    \vspace{-0.25cm}
    \label{fig:auc}
\end{figure}

\noindent
\textbf{Results}:
Figure \ref{fig:roc} presents ROC curves of different algorithms for a setup with increasing dimension $p$ (with $n$ scaling in proportion).
Given the data sampling procedure, NGL close to be the MLE of the model, and hence, achieves the best ROC.
However it involves $p(p-1)/2$ variables and a computational complexity in $\mathcal{O}(p^3)$.
Conversely, we see that LRCC (involving $p\times k$ parameters and with complexity $\mathcal{O}(pk^2+k^3)$) reaches performance close to the optimal as the dimension $p$ increases, even when we set $k$ as low as $10\%$ of $p$.
We thus confirm that a favorable trade-off in terms of performance versus computational load can be achieved in this context.
The scalability of LRCC is further demonstrated by consistently high AUC scores (greater than $0.9$) across various low-rank settings of $k$, as presented in Table~\ref{tab:fullpage_top}.
The reported results correspond to the optimally chosen values of $\lambda$, so we investigate the sensitivity of LRCC to this parameter in Figure~\ref{fig:auc}.
This figure shows that the performance of LRCC under low-rank constraints is sensitive to the choice of $\lambda$, which requires a careful tuning.
In particular, lowest rank settings ($k=0.1p$) tend to require larger values of $\lambda$, whereas the full-rank case performs best with smaller $\lambda$.
For intermediate ranks ($k = 0.35p$, $0.5p$, and $0.75p$), the optimal performance is achieved with moderately adjusted $\lambda$ values.
These findings suggest that $\lambda$ plays a compensatory role in enhancing the performance of LRCC under low-rank constraints.

\begin{table*}[t]  
    \centering
    \begin{tabular}{|c|c|c|c|c|c|c|c|c|c|c|}
        \hline
        n &  p &  $k=10\%p$&  $k=15\%p$ &  $k=25\%p$ &  $k=35\%p$ &  $k=50\%p$ & $k=60\%p$ &  $k=75\%p$ &  $k=90\%p$ &  $k=p$ \\
        \hline
        $p+5$   & $150$    & $0.97$   & $0.95$ & $0.91$ & $0.93$ & $0.92$ & $0.92$ & $0.93$ & $0.97$ & $0.98$     \\
        $-$   & $200$    & $0.98$   & $0.96$ & $0.92$ & $0.94$ & $0.93$ & $0.94$ & $0.95$ & $0.97$ & $0.98$       \\
        $-$   & $250$    & $0.98$   & $0.97$ & $0.93$ & $0.95$ & $0.94$ & $0.95$ & $0.96$ & $0.98$ & $0.96$       \\
        $-$   & $300$    & $0.98$   & $0.95$ & $0.92$ & $0.95$ & $0.94$ & $0.95$ & $0.96$ & $0.94$ & $0.96$       \\
        \hline
    \end{tabular}
    \caption{The average AUC score of LRCC in low sample support setting $n=p+5$ with various dimension $p$ tested for Barabási-Albert (BA) graph and a variety of rank values $k$ are tested for every dimension $p$.}
    \label{tab:fullpage_top} \vspace{-0.5cm}
\end{table*}
\vspace{-0.5cm}

\subsection{Link prediction on Wireless Sensor Network}
\label{subsec:wireless}

This section apply graph learning methods to data generated by a wireless sensor network ($54$ sensors) deployed at the Intel Berkeley Research Lab \cite{IntelWSN}.\\

\begin{figure}[t!]
\clearpage
    \centering
\includegraphics[width=0.5\textwidth,trim=150 470 110 100, clip]{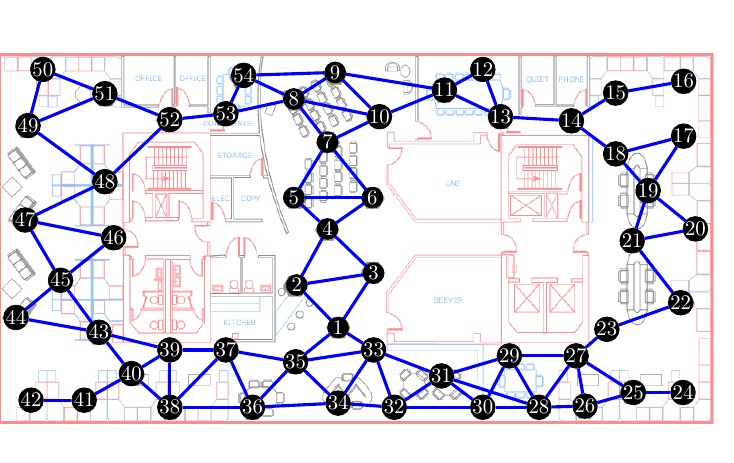}
    \vspace{-1.cm}
    \caption{The connectivity of groundtruth of Intel dataset, established from Eq.~\eqref{eq: rbf}. Background image from \cite{IntelWSN}.} 
\label{fig:connectivity}
\end{figure}

\noindent
\textbf{Setup}: The dataset includes the coordinates of these sensors within the lab, and our objective is to predict the connectivity between them.
Specifically, a reasonable hypothesis for all measurements recorded at each sensor is that a spatial correlation exists, \textit{i.e.}, that sensors that are closer together are more likely to be connected than those farther apart.  
To establish a ground truth based on this principle, a kernel-based function of the form
\begin{equation}
\vspace{-0.15cm}
\label{eq: rbf}
\mathbf{A}_{ij} = \exp\left(-\frac{d(i,j)}{2\gamma^2}\right),
\end{equation}  
is employed \cite{dabush2024}, where \( d(i,j) \) represents the Euclidean distance between sensors \( i \) and \( j \), and \( \gamma \) is the kernel bandwidth, set to 5. A threshold \( \beta = 0.5 \) is applied to determine connectivity:  If \( \mathbf{A}_{ij} \geq \beta \), then \( \mathbf{A}_{ij} = 1 \) (indicating a connection); otherwise, \( \mathbf{A}_{ij} = 0 \) (indicating no connection).  
This ground truth is displayed in Figure~\ref{fig:connectivity}. 
The sensor recorded various measurements, and we focus on the data matrix $ \mathbf{X}^{n \times p} $ constructed using standardized voltage measurements, \( n \) denotes the length of the time series and \( p \) denotes the number of sensors. 
This data matrix is then used to predict connectivity with the different graph learning methods.\\

\begin{figure}[t!]
\clearpage
    \centering
    \includegraphics[width=0.5\textwidth]{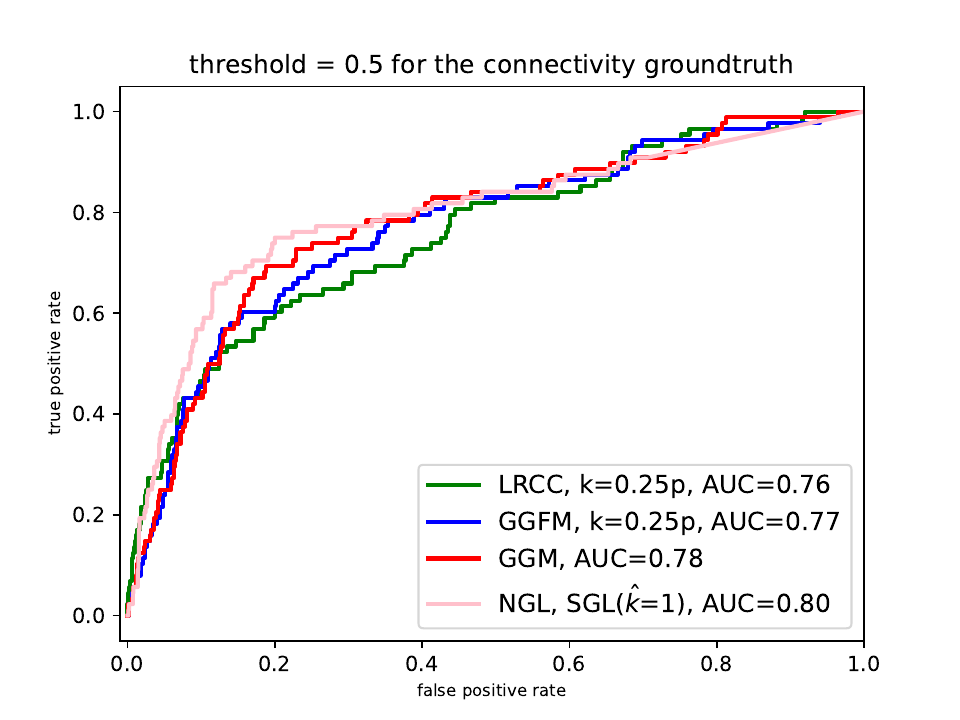}
    \vspace{-0.75cm}
    \caption{\footnotesize The comparison of ROC/AUC for Intel dataset with thres = $0.5$ for the connectivity groundtruth. } 
    \vspace{-0.5cm}
\label{fig:aucIntel}
\end{figure}

\noindent
\textbf{Results}:
Figure~\ref{fig:aucIntel} displays to ROC curve of the different methods.
We observe that NGL reaches again the highest accuracy, which is counterbalanced by its complexity (cf. Section~\ref{sec:Time_Complexity}).
Interestingly, in the most useful zone of low false positive rate range ${\rm PFA}\in[0, 0.1]$, all methods, including LRCC, exhibits performance comparable to NGL.
These observations further underline the favorable trade-off between accuracy and computational efficiency offered by our approach.

\subsection{Link prediction on Animals dataset}
\label{subsec:amimals_data}

This section assesses the clustering capability of LRCC on a semantic dataset.

\noindent
\textbf{Setup}:
We use the \textit{Animals} dataset,\cite{osherson1991default, lake2010discovering, hippert2023learning, kumar2020unified} which contains \( p = 33 \) animals, each described by \( n = 102 \) binary features based on yes/no questions such as ``Has it teeth?'' or ``Is it poisonous?''. These features represent categorical and non-Gaussian data. As no ground truth is available for this dataset, we evaluate the link prediction task using visual inspection: to enhance visualization, each graph node is then clustered into colors using a community detection algorithm based on label propagation \cite{cordasco2010}.
The graphs generated by the methods can be easily validated based on their ability to cluster animals with similar behavioral traits and physical characteristics, which is straightforward and does not require readers to have any domain-specific knowledge.\\

\begin{figure*}[t!]
\clearpage
\centering
\begin{minipage}{0.32\textwidth}
        \centering
        \includegraphics[width=\textwidth]{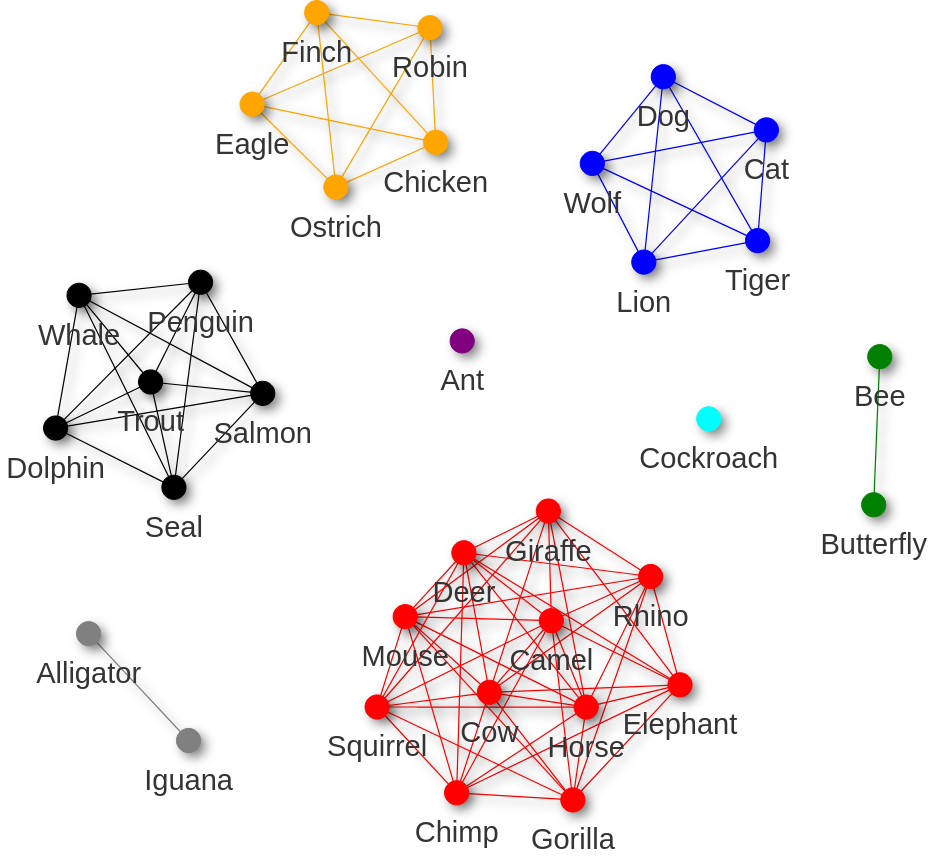}
        \caption*{a) SGL ($\hat{k}=8$) \cite{kumar2020unified}}
    \end{minipage}
    \hfill
    \begin{minipage}{0.32\textwidth}
        \centering
        \includegraphics[width=\textwidth]{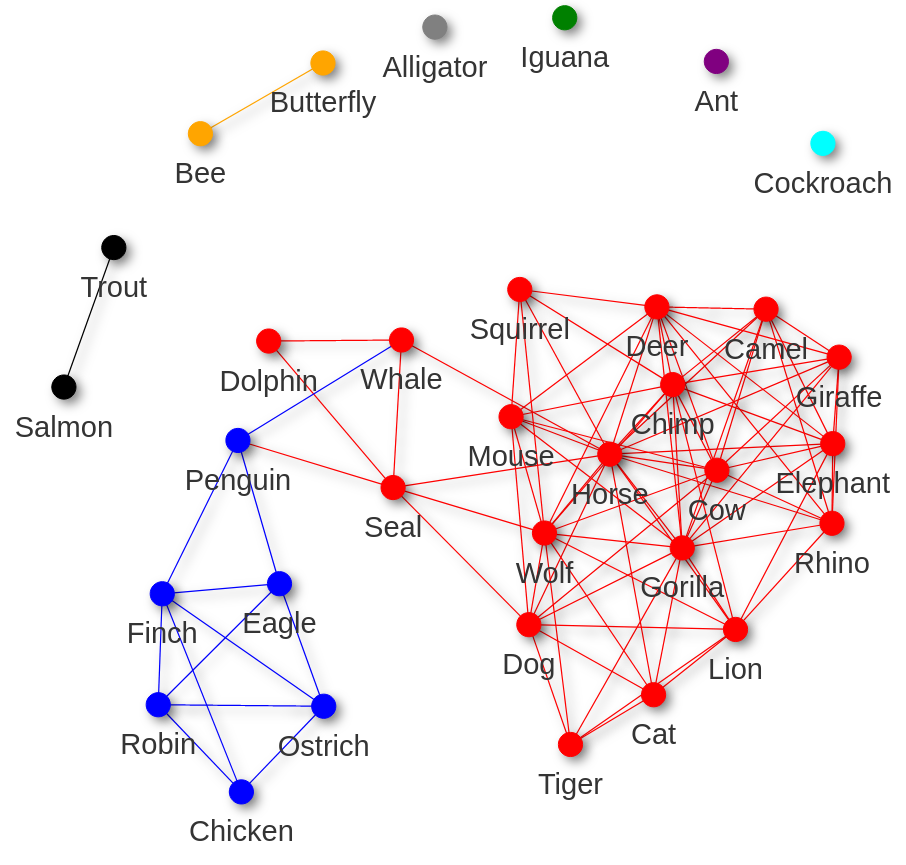}
        \caption*{b) GGM \cite{hippert2023learning}}
    \end{minipage}
    \hfill
    \begin{minipage}{0.32\textwidth}
        \centering
        \includegraphics[width=\textwidth]{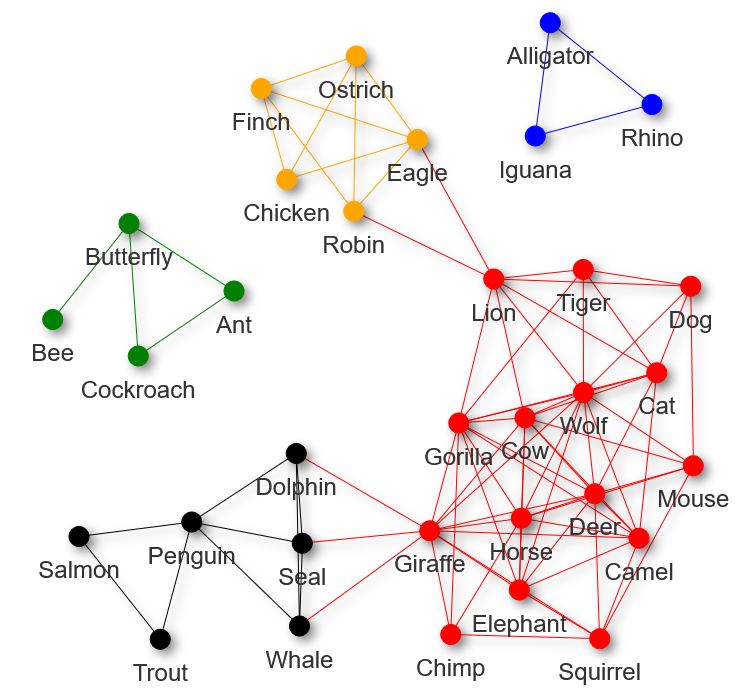}
        \caption*{e) LRCC ($k=12$)}
    \end{minipage}

   
    \caption{Clustering capability example on \textit{Animals} dataset.}
    \label{fig:animal}
\end{figure*}

\noindent
\textbf{Results}: 
Figure~\ref{fig:animal} displays the graphs obtained by SGL (extension of SGL that allows for fixing the number of components in the graph \cite{kumar2020unified}), GGM (standard baseline), and LRCC.
In this setup, SGL was already shown to provide interesting and interpretable results \cite{kumar2020unified, hippert2023learning}, so we focus on this baseline for comparison.
However the method is still of high computational complexity, and requires the user to fix the number of clusters a priori.
Comparatively, we see that GGM yields a graph with only two large clusters and some isolated nodes.
We observe that LRCC achieves an interesting improvement, even when reducing the rank down to a value of $k = 12$ : the graph inferred by LRCC appears well-structured, with no isolated nodes, and reasonable semantic clusters.
These results further support the interest of the proposed method in terms of model dimension-versus-performance trade-off.
%
%
%
%

\section{Conclusion}

In this paper, we introduced a Riemannian optimization based framework for efficient graph learning.
Our approach uses Riemannian optimization to represent the precision matrix with a low-rank conditional correlation structure.
This low-rank formulation significantly reduces computational complexity, making the method scalable to high-dimensional data, in contrast to standard Gaussian Graphical Models (GGMs) and Laplacian learning methods. 
Through experiments on both synthetic and real datasets, we show that our method achieves a strong balance between model performance and computational efficiency.

\bibliographystyle{IEEEtran} 
\bibliography{references} 

\end{document}